\documentclass[conference]{IEEEtran}
\IEEEoverridecommandlockouts
\usepackage{cite}
\usepackage{amsmath,amssymb,amsfonts}
\usepackage{algorithmic}
\usepackage{graphicx}
\usepackage{textcomp}
\usepackage{xcolor}
\usepackage{marvosym}
\usepackage{hyperref}
\usepackage{enumitem}
\usepackage{multirow}
\hypersetup{
  colorlinks=true,
  linkcolor=blue,   
  citecolor=green,  
  urlcolor=red     
}

\def\BibTeX{{\rm B\kern-.05em{\sc i\kern-.025em b}\kern-.08em
    T\kern-.1667em\lower.7ex\hbox{E}\kern-.125emX}}
\begin{document}

\title{Prompt-Guided Generation of Structured Chest X-Ray Report Using a Pre-trained LLM\\

\thanks{\textsuperscript{\Letter}Corresponding authors.}
}

\author{\IEEEauthorblockN{1\textsuperscript{st} Hongzhao Li}
\IEEEauthorblockA{\textit{Network and Data Center} \\
\textit{Northwest University}\\
Xi'an, China \\
lihongzhao@stumail.nwu.edu.cn}
\and
\IEEEauthorblockN{2\textsuperscript{nd} Hongyu Wang}
\IEEEauthorblockA{\textit{School of Computer Science and Technology} \\
\textit{Xi’an University of Posts and Telecommunications}\\
Xi'an, China \\
hywang@xupt.edu.cn
}
\and
\IEEEauthorblockN{3\textsuperscript{rd} Xia Sun\textsuperscript{\Letter}}
\IEEEauthorblockA{\textit{School of Information Science and Technology} \\
\textit{Northwest University}\\
Xi'an, China \\
raindy@nwu.edu.cn}
\and
\IEEEauthorblockN{4\textsuperscript{th} Hua He}
\IEEEauthorblockA{\textit{School of Foreign Languages} \\
\textit{Northwest University}\\
Xi'an, China \\
hehua@nwu.edu.cn
}
\and
\IEEEauthorblockN{5\textsuperscript{th} Jun Feng\textsuperscript{\Letter}}
\IEEEauthorblockA{\textit{School of Information Science and Technology} \\
\textit{Northwest University}\\
Xi'an, China \\
fengjun@nwu.edu.cn}
}

\maketitle

\begin{abstract}
Medical report generation automates radiology descriptions from images, easing the burden on physicians and minimizing errors. However, current methods lack structured outputs and physician interactivity for clear, clinically relevant reports. Our method introduces a prompt-guided approach to generate structured chest X-ray reports using a pre-trained large language model (LLM). First, we identify anatomical regions in chest X-rays to generate focused sentences that center on key visual elements, thereby establishing a structured report foundation with anatomy-based sentences. We also convert the detected anatomy into textual prompts conveying anatomical comprehension to the LLM. Additionally, the clinical context prompts guide the LLM to emphasize interactivity and clinical requirements. By integrating anatomy-focused sentences and anatomy/clinical prompts, the pre-trained LLM can generate structured chest X-ray reports tailored to prompted anatomical regions and clinical contexts. We evaluate using language generation and clinical effectiveness metrics, demonstrating strong performance.
\end{abstract}


\section{Introduction}
\label{sec:intro}
Medical report generation aims to automatically create textual descriptions from radiological images like chest X-rays, a time-consuming and error-prone task even for experienced radiologists, while demand exceeds current medical capabilities \cite{ganeshan2018structured}. Automatically generating reports could reduce physician workload and diagnostic errors. This field integrates computer vision and natural language processing areas, making it an important area of multimodal research \cite{otter2020survey}.

Previous work on medical report generation has primarily employed encoder-decoder models \cite{chen2020generating}, using CNN visual encoders to extract image features and Transformer \cite{vaswani2017attention} text decoders to convert those features into textual output. Research has focused on the end-to-end goal of generating reports from radiological images. However, some key issues have been overlooked. \textbf{Structural Deficiency:} Unstructured radiology reports reduce clarity and make it challenging for physicians to identify critical information \cite{ganeshan2018structured}. However, current end-to-end methods are trained on datasets containing unstandardized reports from diverse sources, hindering the generation of uniformly structured outputs. This poses a barrier to optimal patient care, and yet there is limited current work addressing this lack of structure.
\textbf{Lack of Interpretability and Interactivity:} Existing methods often insufficiently explain reasoning through local attention visualizations rather than clearly elucidating decisions. Furthermore, doctors cannot effectively intervene in the generation process to adjust content based on patient context. This opacity and rigidity restrict practical application \cite{miller2019explanation}.

\begin{figure}
\centering
\includegraphics[width=0.95\linewidth]{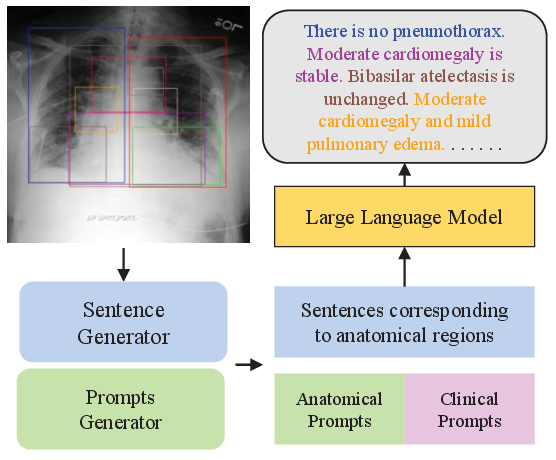}
\caption{The high-level representation of our architecture. Each sentence can be explicitly associated with a specific region.}
\label{fig1}
\vspace{-0.5cm}
\end{figure}

\begin{figure*}
\centering
\includegraphics[width=0.9\linewidth]{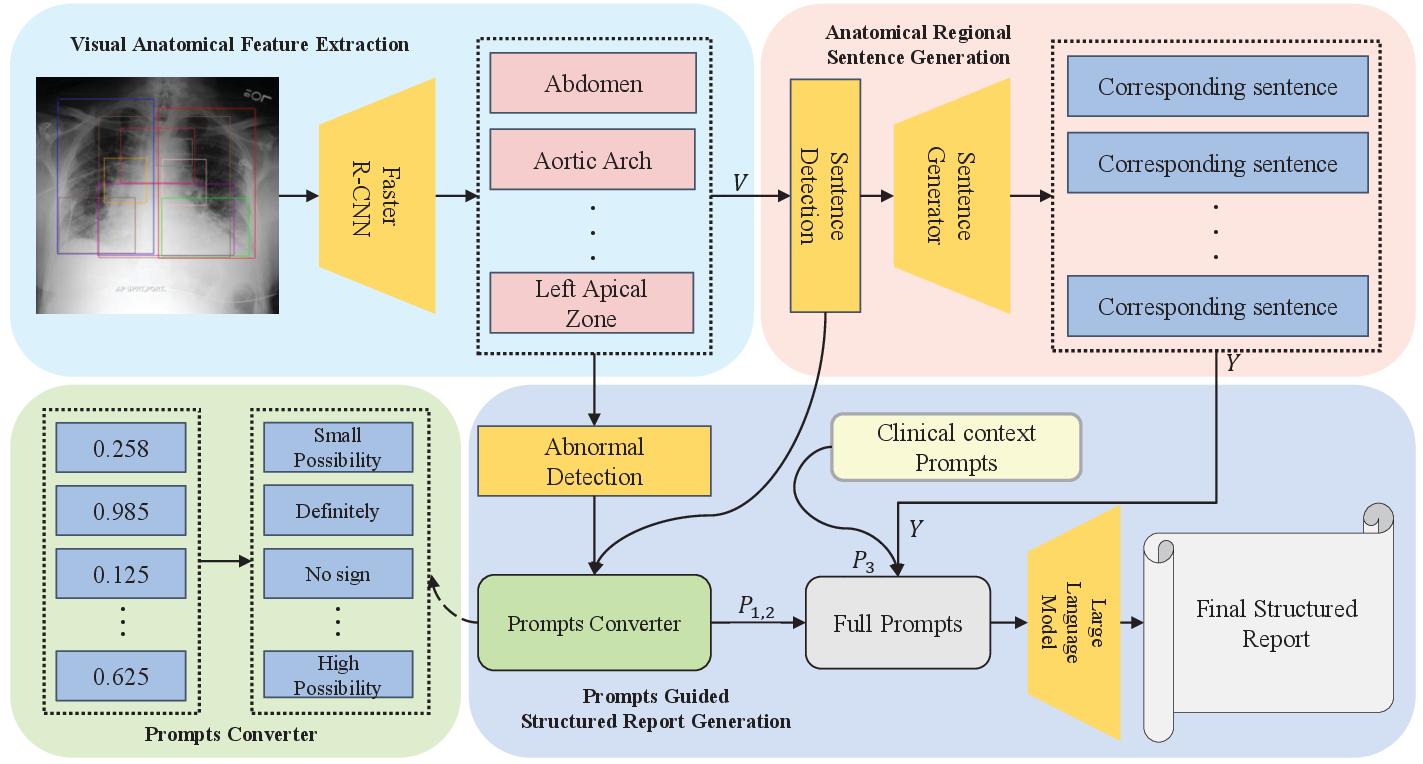}
\caption{In the architecture overview, the process includes identifying and extracting anatomical region features, generating region descriptions, and ultimately integrating anatomical prompts with clinical context prompts to produce a structured report.}
\label{fig2}
\end{figure*}
In response to the aforementioned issues, we propose an structured report generation model (Fig.~\ref{fig1}) using a pre-trained large language model (LLM) guided by anatomical regions and clinical contextual prompts to achieve high interpretability and interactivity. \textbf{First,} we detect anatomical regions in chest X-rays to generate region-focused descriptions, establishing an anatomy foundation for structured report. We additionally translate anatomical information into textual prompts, enabling subsequent modules comprehension of anatomy-guided data for more accurate descriptions. \textbf{Second,} our model incorporates clinical contextual information, including the patient's medical history and the reason for examination, etc., typically provided by the physician. This interactivity allows physicians to actively participate in report generation by providing relevant information. \textbf{Lastly,} we utilize a large language model to integrate anatomical region descriptions, anatomical prompts, and clincial contextual prompts into a single anatomically based structured by coordinating and consolidating these data sources.

Overall, our work contributes as follows: 

\begin{itemize}
\item We introduced an anatomy-guided structured report generation framework by identifying anatomical regions to construct anatomy-centric sentences, establishing the basis for structured reports with lucid expressions. We also integrated patient contextual information for comprehensive clinical understanding.

\item By inputting anatomy-focused sentences and anatomy/clinica prompts into a large language model, we produced structured, interpretable reports bearing anatomical and clinical relevance. Additionally, our architecture empowers physicians to provide clinical context, enabling intervention and adjustment in report generation to address variable clinical needs.

\item We evaluated our method on the MIMIC-CXR dataset \cite{johnson2019mimic} using both natural language generation and clinical effectiveness metrics. We performed well on both sets of metrics and conducted detailed visual experiments to demonstrate and support the structured nature of our generated reports, as well as the interpretability and interactivity of the method.
\end{itemize}

\section{Related work}
\noindent \textbf{Medical report generation.} Recent frameworks build on medical image captioning to incorporate expert or external knowledge for richer contextual understanding in generated radiology reports \cite{huang2023kiut,yang2023radiology,wang2023metransformer}. Recent approaches like the MET \cite{wang2023metransformer}, Kiut \cite{wang2023metransformer}, and KBMA \cite{yang2023radiology} incorporate specialized knowledge through fusion of expert tokens, multimodal injection, and trainable knowledge bases to enhance contextual understanding in radiology report generation. The work \cite{tanida2023interactive} uses anatomical regions for the first time to generate reports. To further increase interpretability, some efforts introduce memory networks \cite{chen2020generating,chen2022cross,qin2022reinforced} to store knowledge and learn implicit image-text links. However, unlike these algorithmically synthesized components imitating expertise, we use actual clinical documents to provide contextual knowledge. Furthermore, we guide the generation process using anatomical regions, enhancing interpretability by specifying clear sentence-to-region mappings within the report.

\noindent \textbf{Large Language Model.} 
Recent advances in Transformer \cite{vaswani2017attention} and computing have enabled very large language models (LLMs) like ChatGPT \cite{openai2023gpt4} with enhanced performance on text generation and translation. Architectures like CLIP \cite{radford2021learning} achieve multimodal understanding via image-text pretraining. Additionally, recent work guides LLMs via prompts rather than explicit training. For example, VisualChatGPT \cite{wu2023visual} connects ChatGPT with vision models for image-inclusive conversations, exemplifying this more flexible prompt-based approach to unleashing LLM potential. The idea of guiding LLMs through prompts has inspired our research. Guiding an LLM through anatomical region prompts results in structured, interpretable reports. Furthermore, clinical context prompts facilitate physician interactivity, enhancing the practical utility of the generated reports.

\section{Methodology}
We propose a structured report generation framework, guided by anatomy and clinical prompts, to simulate radiologist workflow. First, we identify anatomical regions in chest X-rays and extract per-region features \hyperref[section:3.1]{(Section 3.1)}. Then, a sentence generator produces region descriptions, forming the basis for the structured report \hyperref[section:3.2]{(Section 3.2)}. Concurrently, we generate anatomical prompts indicating sentence presence and abnormalities per region \hyperref[section:3.3]{(Section 3.3)}. Finally, we integrate region descriptions, anatomical prompts, and clinical context from doctors into prompts for a large language model, it generates the final structured report \hyperref[section:3.4]{(Section 3.4)}. We show the proposed architecture in (Fig.~\ref{fig2}) and describe these steps in more detail below.

\subsection{Anatomy Region Detection and Feature Extraction}
\label{section:3.1}
We employed Faster R-CNN \cite{ren2015faster} with a ResNet-50 backbone \cite{he2016deep} for anatomy detection and feature extraction. Faster R-CNN generates region proposals, then extracts features via RoI pooling and classifies anatomy regions, optimized using the standard Faster R-CNN loss. Next, for each detected region, we pooled the features and transformed them into a 1024-dimensional representation as the image feature:

\begin{equation}
V=\operatorname{Faster\ R-CNN}\left(Img\right).
\end{equation}

Finally, it outputs 29 anatomy regions along with visual features \(V\in \mathbb{R}^{29 \times 1024}\) capturing morphological and pathological information per region.

\subsection{Sentence Generator}
\label{section:3.2}
To generate region sentences, we employed a Transformer Decoder model similar to \cite{radford2019language}, which attends to context from preceding tokens. We integrated regional visual features into the attention computation, allowing the model to jointly consider previous tokens along with anatomy visuals during text generation. We trained the model by minimizing cross-entropy loss to align the generated text with expected reports. Ultimately, this facilitates learning language expressions for anatomical regions, forming the basis of our structured report.
\begin{equation}
Y=\operatorname{SentenceGen}\left(V\right).
\end{equation}
Specifically, \(Y\in \mathbb{R}^{29 \times l}\) is the set of generated sentences for the 29 anatomy regions that together form the structured report foundation, where \(l\) represents the sentence length.
\begin{figure}
\centering
\includegraphics[width=\linewidth]{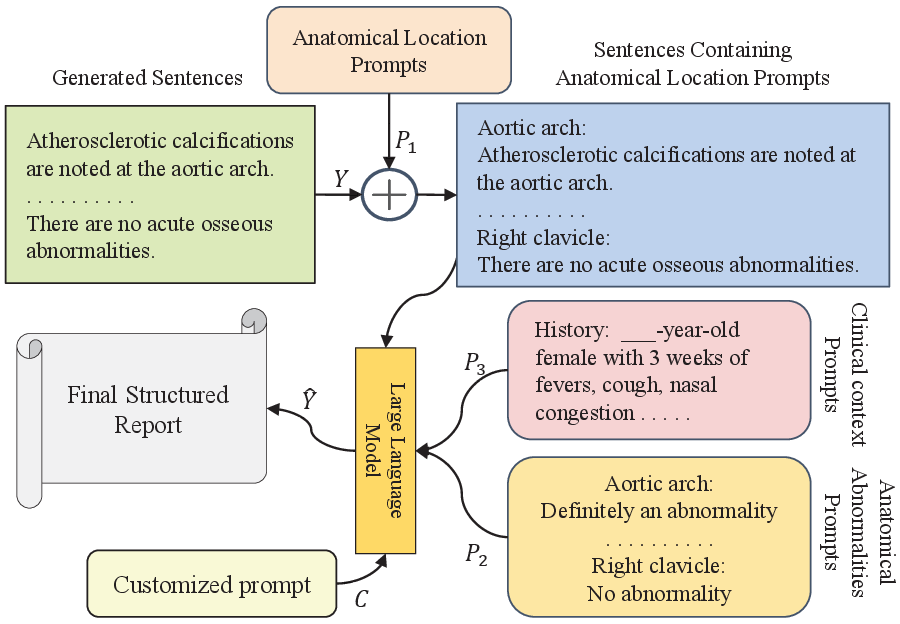}
\caption{An example is presented to illustrate the inputs for the LLM.}
\label{fig3}
\end{figure}

\begin{table*}
\caption{\label{tab1}Comparison results on MIMIC-CXR show that the numbers in bold represent the best results per column.}
\setlength{\tabcolsep}{1mm}
\centering
\begin{tabular}{cc@{\hspace{3mm}}|cccccc|ccc}
\hline
Method & Year & BLEU-1 & BLEU-2 & BLEU-3 & BLEU-4 & METEOR & ROUGE-L & F1 & Precision & Recall\\
\hline
R2GEN \cite{chen2020generating} & 2020 & 0.353 & 0.218 & 0.145 & 0.103 & 0.142 & 0.270 & 0.276 & 0.333 & 0.273\\
R2GENCMN \cite{chen2022cross} & 2022 & 0.353 & 0.218 & 0.148 & 0.106 & 0.149 & 0.284 & 0.278 & 0.334 & 0.275\\
CMM+RL \cite{qin2022reinforced} & 2022 & 0.381 & 0.232 & 0.155 & 0.109 & 0.151 & 0.287 & 0.292 & 0.342 & 0.294\\
CvT \cite{nicolson2023improving} & 2023 & 0.392 & 0.245 & 0.169 & 0.124 & 0.153 & 0.285 & 0.384 & 0.359 & 0.412  \\
MET \cite{wang2023metransformer} & 2023 & 0.386 & 0.250 & 0.169 & 0.124 & 0.152 & \textbf{0.291} & 0.311 & 0.364 & 0.309\\
KiUT \cite{huang2023kiut} & 2023 & 0.393 & 0.243 & 0.159 & 0.113 & 0.160 & 0.285 & 0.321 & 0.371 & 0.318\\
KBMA \cite{yang2023radiology} & 2023 & 0.386 & 0.237 & 0.157 & 0.111 & - & 0.274 & 0.352 & 0.420 & 0.339\\
\hline
Ours & 2023 & \textbf{0.395} & \textbf{0.260} & \textbf{0.178} & \textbf{0.131} & \textbf{0.161} & 0.261 & \textbf{0.441} & \textbf{0.469} & \textbf{0.470}\\
\hline
\end{tabular}
\end{table*}

\subsection{Anatomy Prompts Generation}
\label{section:3.3}
The Anatomy Prompts Generation module has three components: sentence detection, abnormal detection, and a prompts converter. The first two are binary classifiers indicating whether the region should generate sentence and if abnormalities are present. For example, sentence detection may include the heart region as crucial, while abnormal detection flags heart abnormalities. We optimized with binary cross-entropy loss. Then, the prompts converter translates the classifications into clear prompts for language models, such as "The aortic arch is definitely abnormal".
\begin{equation}
P_{1,2}=Prompt\ Generation\left(V\right),
\end{equation}
where \(P_{1}\) and \(P_{2}\) are the anatomical location and abnormality prompts. Ultimately, this enables effective anatomical prompt incorporation to integrate anatomy into decision-making, supporting structured report generation.

\subsection{Structured report generation}
\label{section:3.4}
Large language models (LLMs) possess robust medical knowledge and reasoning abilities. Appropriate prompts allow them to produce precise medical reports. When anatomical descriptions and prompts are organized, we integrate them with clinical context \(P_3\) from doctors into LLM (we used GPT-4 \cite{openai2023gpt4}). Specifically, \(P_3\) includes the history, indications, reasons for examination provided by the physician contained in the dataset. This provides an opportunity for the physician to participate in the interaction. We also designed simple custom prompts \(C\) (e.g. "Generate a structured report based on the anatomical and clinical details"). This enables the LLM to generate appropriately structured outputs.
\begin{equation}
\hat{Y}=\text{LLM}(C,Y,P_1,P_2,P_3)
\end{equation}
In the end, the LLM synthesizes the sentence descriptions \(Y\), anatomical prompts \(P_{1,2}\), and patient context \(P_3\) into a comprehensive, clinically practical structured report \(\hat{Y}\) (Fig.~\ref{fig3}) .

\section{Experiments}
\subsection{Datasets and Metrics}
The recently released MIMIC-CXR \cite{johnson2019mimic} is currently the largest public dataset containing many chest radiograph images and reports. In total, this dataset has 377,110 images and 227,835 reports for 64,588 patients. For experimental and fair comparisons, we followed the previous methodology \cite{chen2020generating,huang2023kiut,yang2023radiology,wang2023metransformer,chen2022cross,qin2022reinforced} used the official MIMIC-CXR divisions: 222,758 samples for training, 1,808 for validation, and 3,269 for testing. Furthermore, we utilized labels generated by Chest ImaGenome \cite{wu2021chest}, where the tags succinctly represent the 29 chest anatomical regions in the image, aligning with the sentences describing each region in the report.

We evaluated radiology report generation using standard natural language generation (NLG) metrics and a clinical efficiency (CE) metric. The NLG metrics were BLEU \cite{papineni2002bleu}, METEOR \cite{banerjee2005meteor}, and ROUGE \cite{lin2004rouge} scores, which are standard metrics used to assess the fluency of generated natural language. As NLG metrics insufficiently measure clinical correctness, the CE metric utilized 14 common disease type labels to compute F1, precision and recall versus basic facts and reports, thereby quantitatively measuring clinical correctness.
\begin{table}
\caption{\label{tab2}Six prominent regions are identified as follows: right lung (RL), left lung (LL), spine (SP), mediastinum (MED), cardiac silhouette (CS), and abdomen (AB).}
\setlength{\tabcolsep}{1mm}
\centering
\begin{tabular}{ccccccc}
\hline
Region & RL & LL & SP & MED & CS & AB\\
\hline
IoU & 0.891 & 0.900 & 0.943 & 0.827 & 0.782 & 0.896  \\
METEOR & 0.121 & 0.113 & 0.173 & 0.123 & 0.101 & 0.109  \\
\hline
\end{tabular}
\end{table}
\subsection{Implementation Details}
For anatomy detection, we did not use the Faster R-CNN \cite{ren2015faster} features directly to avoid over-coupling. Instead, we pooled and transformed the region features, ensuring detection performance. We extracted 29 regions and 1024-dimensional visual features as inputs for generation. The classifiers use three FC layers (1024-512-128-1) with ReLU activations for non-linearity. The sentence generator has three 8-headed attention layers with 512 units each. We trained all modules on one NVIDIA 3090 GPU over three stages: first, train anatomy detection; then add and train the two classifiers; finally, add and train the sentence generator. Importantly, each new module trained concurrently with the previously trained ones to maintain performance. The integration module uses fixed GPT-4 without separate training. All modules used mixed precision, AdamW, learning rate decay, and early stopping.

\begin{table}
\caption{\label{tab3}Results from sentence detection and abnormality detection.}
\centering
\begin{tabular}{ccccc}
\hline
Module & Regions & F1 & P & R\\
\hline
\multirow{3}{*}{Sentence Detection} & All & 0.701 & 0.571 & 0.906  \\
& Abnormal & 0.965 & 1.000 & 0.932 \\
& Normal & 0.489 & 0.341 & 0.869 \\
\hline
Abnormal Detection & All & 0.557 & 0.394 & 0.951   \\
\hline
\end{tabular}
\end{table}

\begin{figure*}[h]
\centering
\includegraphics[width=\linewidth]{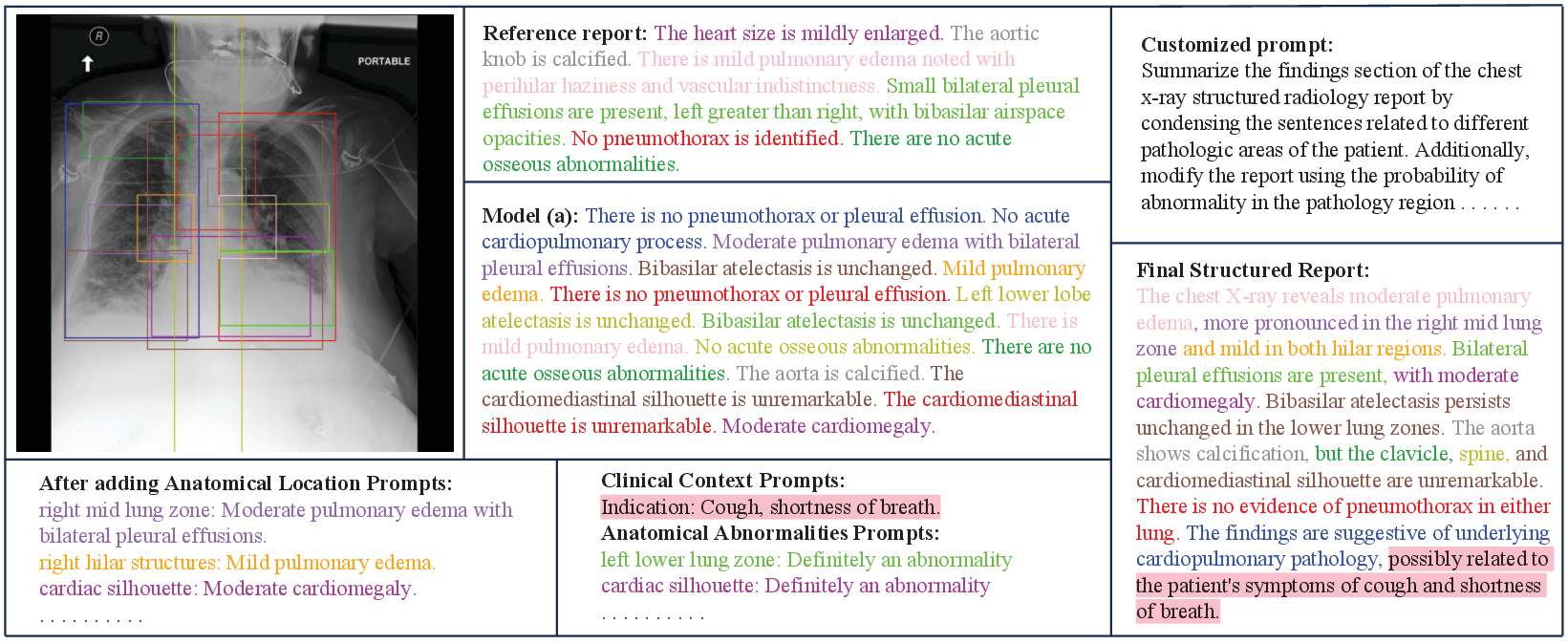}
\caption{We represented the detected anatomical regions, the corresponding generated sentences, and the semantically matched reference sentences using the same color, while also highlighting the clinical context.}
\label{fig4}
\end{figure*}

\subsection{Quantitative results}
\noindent \textbf{Comparison with state-of-the-art.} We conducted a comprehensive comparison of our model with recent methods, as referenced in \cite{chen2020generating,huang2023kiut,yang2023radiology,wang2023metransformer,chen2022cross,qin2022reinforced}, using NLG and CE metrics (Tab.~\ref{tab1}). These methods encompass a diverse array of techniques, including the utilization of knowledge graphs, large language models, reinforcement learning, and various network architectures. In terms of NLG metrics, our model emerges as the top performer overall, displaying remarkable language quality and fluency. Notably, we observe a minor decrease only in ROUGE-L, which doesn't compromise the model's excellence in report generation. Importantly, our model outperforms others significantly in clinical correctness. Our reports not only exhibit superior language quality but also align more closely with real diagnoses. This outstanding performance is evident across all CE metrics, underscoring the exceptional medical accuracy and reliability of our approach.

\noindent \textbf{Analysis of Key Regions.} In Tab.~\ref{tab2}, we present the IoU and METEOR scores for six key regions, where IoU gauges localization accuracy, and METEOR assesses semantic report consistency. Across most regions, high IoU values indicate precise identification. And SP and MED exhibit higher METEOR scores, implying more consistent reports. However, RL and LL show a certain degree of negative correlation, indicating that after imperfect detection, subsequent modules can still accurately convey anatomical semantics. Therefore, minor deviations in detection will not have a decisive impact on the final report.

\begin{table}
    \caption{\label{tab4}The displayed results depict the outcomes of the ablation experiments. NLG scores represent the average of all metrics in NLG. In this context, SDet corresponds to sentence detection, ADet corresponds to anomaly detection, loss is represented by \(L\), and prompt is indicated by \(P\). }
    \centering
    \setlength{\tabcolsep}{1.4mm}
    \begin{tabular}{cccccc|cccc}
        \hline
        \multirow{2}{*}{Model} & \multicolumn{2}{c}{SDet} & \multicolumn{2}{c}{ADet} & \multirow{2}{*}{\(P_{3}\)} & \multirow{2}{*}{NLG} & \multicolumn{3}{c}{CE}  \\
        & \(L_{1}\) & \(P_{1}\) & \(L_{2}\) & \(P_{2}\) & & & F1 & P & R \\
        \hline
(a) & & & & & & 0.183 & 0.304 & 0.355 & 0.301 \\
(b) & \(\checkmark\) & & \(\checkmark\) & &  & \textbf{0.233} & 0.404 & 0.432 & 0.432 \\
(c) & \(\checkmark\) & \(\checkmark\) & \(\checkmark\) & &  & 0.229 & 0.405 & 0.421 & 0.450 \\
(d) & \(\checkmark\)& & \(\checkmark\) & \(\checkmark\) &  & 0.232 & 0.434 & 0.450 & 0.459 \\
(e) & \(\checkmark\) & & \(\checkmark\) & & \(\checkmark\) & \textbf{0.233} & 0.408 & 0.436 & 0.437 \\
(f) & \(\checkmark\) & \(\checkmark\) & \(\checkmark\) & \(\checkmark\) & \(\checkmark\) & 0.231 & \textbf{0.441} & \textbf{0.469} & \textbf{0.470} \\
        \hline
    \end{tabular}
\end{table}
\noindent \textbf{Evaluation of Sentence and Anomaly Detection.} Tab.~\ref{tab3} shows results for the sentence and anomaly detection modules. The sentence detection module focuses on whether the region generates a sentence, while the abnormal detection module focuses on whether the region is abnormal. For sentence detection, our model demonstrates a commendable overall high recall, indicating a propensity to generate a larger number of sentences. Nevertheless, the precision of normal sentences is relatively low, resulting in the inclusion of more additional normal sentences in the generated reports. This expanded sentence generation contributes to increased textual detail, and consequently, our model's ROUGE-L score is marginally underrepresented. However, it's crucial to highlight that the decision to include normal sentences is highly subjective in a clinical context and depends on the physician's judgment. Some physicians may choose not to mention normal areas. Therefore, the reduction in normal accuracy is comprehensible. In contrast, the accuracy in detecting abnormalities is consistently at 1.0, given that abnormal regions are inherently included in the reference reports. In terms of abnormal detection, we found a higher recall rate but a lower precision rate. A higher recall rate can effectively identify most true abnormalities. In a clinical setting, emphasizing achieving high sensitivity and ensuring timely detection is crucial, highlighting the importance of high recall in practical applications.

\noindent \textbf{Ablation Analysis.} Tab.~\ref{tab4} presents ablation studies focusing on loss functions and prompts. We employ a unique custom prompt \(C\) in all models except for model (a), enabling the LLM to eliminate redundant sentences and make more effective use of the generated information. In the case of model (b), the inclusion of sentence and anomaly detection losses proves to be instrumental in significantly enhancing performance, guiding the model's attention towards abnormalities even in the absence of explicit prompts. Upon scrutinizing prompt analyses, distinct effects emerge. Anatomical and contextual prompts (Models c-e) exert a positive influence on clinical accuracy, underscoring their pivotal role. However, it's noteworthy that they marginally decrease NLG metrics. This decrease can be attributed to the generated reports containing more anatomical details than reference report. Importantly, this doesn't imply a reduction in practicality; instead, it suggests that the report is more structured, comprehensive, and consistent with medical standards.

\subsection{Qualitative results}
In Fig.\ref{fig4}, we visually illustrate the generation process for a specific example. Our visualization also enables demonstrating the interpretability, interactivity, and structural nature of the report. First, we observed Model (a), which represents sentences generated before prompts inclusion. The colors of anatomical regions boxes correspond to generated sentences colors, strengthening interpretability as physicians can associate sentences with anatomical regions. While Model (a) laid the initial foundation for an anatomy-oriented structured report, the sentences lacked specific anatomical information and clinical context while redundant. To address this, location, abnormality, context and customized prompts were incorporated, whereby the LLM successfully generated more accurate, detailed and structured reports. The final report focuses on anatomical areas, providing region-specific findings regarding pulmonary edema severity across lung areas and noting bilateral pleural effusion. Additionally, cardiac enlargement and aortic calcification are described. This structured report boosts understanding and comparison by supplying region-specific anatomical and examination details alongside abnormality presence/extent descriptions. Moreover, physician-furnished contextual information like indications is included, relating reported cough, shortness of breath to suggested cardiopulmonary issues. This supplementary context enables fuller clinical condition comprehension while providing diagnosis/treatment clues, demonstrating prompts can align model focus with expectations.

\section{Conclusion}
We propose an structured report generation model using a pre-trained large language model LLM guided by anatomical regions and clinical contextual prompts to achieve high interpretability and interactivity. First, we introduced anatomical structure detection to establish structured, anatomy-centric visual descriptions, a novel contribution. Second, through interactive textual prompts and a large language model, we enabled physician guidance catered to variable clinical contexts, also lacking in prior work. Our emphasis on report structure and process interpretability and interactivity, coupled with strong performance on relevant metrics, contributes to the solution of pervasive limitations in this domain.

\section*{Acknowledgment}
This research was supported by the National Natural Science Foundation of China (NSFC Grant No. 62073260, 62001380); Key \& Dprojects in Shaanxi Province (No. 2023-YBSF-493).

\bibliographystyle{IEEEtran}

\bibliography{icme2024template}

\begin{thebibliography}{10}
\providecommand{\url}[1]{#1}
\csname url@samestyle\endcsname
\providecommand{\newblock}{\relax}
\providecommand{\bibinfo}[2]{#2}
\providecommand{\BIBentrySTDinterwordspacing}{\spaceskip=0pt\relax}
\providecommand{\BIBentryALTinterwordstretchfactor}{4}
\providecommand{\BIBentryALTinterwordspacing}{\spaceskip=\fontdimen2\font plus
\BIBentryALTinterwordstretchfactor\fontdimen3\font minus \fontdimen4\font\relax}
\providecommand{\BIBforeignlanguage}[2]{{%
\expandafter\ifx\csname l@#1\endcsname\relax
\typeout{** WARNING: IEEEtran.bst: No hyphenation pattern has been}%
\typeout{** loaded for the language `#1'. Using the pattern for}%
\typeout{** the default language instead.}%
\else
\language=\csname l@#1\endcsname
\fi
#2}}
\providecommand{\BIBdecl}{\relax}
\BIBdecl

\bibitem{ganeshan2018structured}
D.~Ganeshan, P.-A.~T. Duong, L.~Probyn, L.~Lenchik, T.~A. McArthur, M.~Retrouvey, E.~H. Ghobadi, S.~L. Desouches, D.~Pastel, and I.~R. Francis, ``Structured reporting in radiology,'' \emph{Academic radiology}, vol.~25, no.~1, pp. 66--73, 2018.

\bibitem{otter2020survey}
D.~W. Otter, J.~R. Medina, and J.~K. Kalita, ``A survey of the usages of deep learning for natural language processing,'' \emph{IEEE transactions on neural networks and learning systems}, vol.~32, no.~2, pp. 604--624, 2020.

\bibitem{chen2020generating}
Z.~Chen, Y.~Song, T.-H. Chang, and X.~Wan, ``Generating radiology reports via memory-driven transformer,'' \emph{arXiv preprint arXiv:2010.16056}, 2020.

\bibitem{vaswani2017attention}
A.~Vaswani, N.~Shazeer, N.~Parmar, J.~Uszkoreit, L.~Jones, A.~N. Gomez, {\L}.~Kaiser, and I.~Polosukhin, ``Attention is all you need,'' \emph{Advances in neural information processing systems}, vol.~30, 2017.

\bibitem{miller2019explanation}
T.~Miller, ``Explanation in artificial intelligence: Insights from the social sciences,'' \emph{Artificial intelligence}, vol. 267, pp. 1--38, 2019.

\bibitem{johnson2019mimic}
A.~E. Johnson, T.~J. Pollard, N.~R. Greenbaum, M.~P. Lungren, C.-y. Deng, Y.~Peng, Z.~Lu, R.~G. Mark, S.~J. Berkowitz, and S.~Horng, ``Mimic-cxr-jpg, a large publicly available database of labeled chest radiographs,'' \emph{arXiv preprint arXiv:1901.07042}, 2019.

\bibitem{huang2023kiut}
Z.~Huang, X.~Zhang, and S.~Zhang, ``Kiut: Knowledge-injected u-transformer for radiology report generation,'' in \emph{Proceedings of the IEEE/CVF Conference on Computer Vision and Pattern Recognition}, 2023, pp. 19\,809--19\,818.

\bibitem{yang2023radiology}
S.~Yang, X.~Wu, S.~Ge, Z.~Zheng, S.~K. Zhou, and L.~Xiao, ``Radiology report generation with a learned knowledge base and multi-modal alignment,'' \emph{Medical Image Analysis}, vol.~86, p. 102798, 2023.

\bibitem{wang2023metransformer}
Z.~Wang, L.~Liu, L.~Wang, and L.~Zhou, ``Metransformer: Radiology report generation by transformer with multiple learnable expert tokens,'' in \emph{Proceedings of the IEEE/CVF Conference on Computer Vision and Pattern Recognition}, 2023, pp. 11\,558--11\,567.

\bibitem{tanida2023interactive}
T.~Tanida, P.~M{\"u}ller, G.~Kaissis, and D.~Rueckert, ``Interactive and explainable region-guided radiology report generation,'' in \emph{Proceedings of the IEEE/CVF Conference on Computer Vision and Pattern Recognition}, 2023, pp. 7433--7442.

\bibitem{chen2022cross}
Z.~Chen, Y.~Shen, Y.~Song, and X.~Wan, ``Cross-modal memory networks for radiology report generation,'' \emph{arXiv preprint arXiv:2204.13258}, 2022.

\bibitem{qin2022reinforced}
H.~Qin and Y.~Song, ``Reinforced cross-modal alignment for radiology report generation,'' in \emph{Findings of the Association for Computational Linguistics: ACL 2022}, 2022, pp. 448--458.

\bibitem{openai2023gpt4}
OpenAI, ``Gpt-4 technical report,'' 2023.

\bibitem{radford2021learning}
A.~Radford, J.~W. Kim, C.~Hallacy, A.~Ramesh, G.~Goh, S.~Agarwal, G.~Sastry, A.~Askell, P.~Mishkin, J.~Clark \emph{et~al.}, ``Learning transferable visual models from natural language supervision,'' in \emph{International conference on machine learning}.\hskip 1em plus 0.5em minus 0.4em\relax PMLR, 2021, pp. 8748--8763.

\bibitem{wu2023visual}
C.~Wu, S.~Yin, W.~Qi, X.~Wang, Z.~Tang, and N.~Duan, ``Visual chatgpt: Talking, drawing and editing with visual foundation models,'' \emph{arXiv preprint arXiv:2303.04671}, 2023.

\bibitem{ren2015faster}
S.~Ren, K.~He, R.~Girshick, and J.~Sun, ``Faster r-cnn: Towards real-time object detection with region proposal networks,'' \emph{Advances in neural information processing systems}, vol.~28, 2015.

\bibitem{he2016deep}
K.~He, X.~Zhang, S.~Ren, and J.~Sun, ``Deep residual learning for image recognition,'' in \emph{Proceedings of the IEEE conference on computer vision and pattern recognition}, 2016, pp. 770--778.

\bibitem{radford2019language}
A.~Radford, J.~Wu, R.~Child, D.~Luan, D.~Amodei, I.~Sutskever \emph{et~al.}, ``Language models are unsupervised multitask learners,'' \emph{OpenAI blog}, vol.~1, no.~8, p.~9, 2019.

\bibitem{nicolson2023improving}
A.~Nicolson, J.~Dowling, and B.~Koopman, ``Improving chest x-ray report generation by leveraging warm starting,'' \emph{Artificial Intelligence in Medicine}, vol. 144, p. 102633, 2023.

\bibitem{wu2021chest}
J.~T. Wu, N.~N. Agu, I.~Lourentzou, A.~Sharma, J.~A. Paguio, J.~S. Yao, E.~C. Dee, W.~Mitchell, S.~Kashyap, A.~Giovannini \emph{et~al.}, ``Chest imagenome dataset for clinical reasoning,'' \emph{arXiv preprint arXiv:2108.00316}, 2021.

\bibitem{papineni2002bleu}
K.~Papineni, S.~Roukos, T.~Ward, and W.-J. Zhu, ``Bleu: a method for automatic evaluation of machine translation,'' in \emph{Proceedings of the 40th annual meeting of the Association for Computational Linguistics}, 2002, pp. 311--318.

\bibitem{banerjee2005meteor}
S.~Banerjee and A.~Lavie, ``Meteor: An automatic metric for mt evaluation with improved correlation with human judgments,'' in \emph{Proceedings of the acl workshop on intrinsic and extrinsic evaluation measures for machine translation and/or summarization}, 2005, pp. 65--72.

\bibitem{lin2004rouge}
C.-Y. Lin, ``Rouge: A package for automatic evaluation of summaries,'' in \emph{Text summarization branches out}, 2004, pp. 74--81.

\end{thebibliography}


\end{document}